\documentclass[10pt,twocolumn,letterpaper]{article}

\usepackage{iccv}
\usepackage{times}
\usepackage{epsfig}
\usepackage{graphicx}
\usepackage{amsmath}
\usepackage{amssymb}
\usepackage{tabularx}
\usepackage{footmisc}
\usepackage[pagebackref=true,breaklinks=true,letterpaper=true,colorlinks,bookmarks=false]{hyperref}

\iccvfinalcopy % *** Uncomment this line for the final submission

\def\iccvPaperID{****} % *** Enter the ICCV Paper ID here

\ificcvfinal\fi

\title{1st Place Solution of Multiview Egocentric Hand Tracking Challenge ECCV2024}

\author{Minqiang Zou , Zhi Lv, Riqiang Jin, Tian Zhan, Mochen Yu, Yao Tang, Jiajun Liang\textsuperscript{*}\\
Jiiov Technology\\
{\tt\small{\{minqiang.zou, zhi.lv, riqiang.jin, tian.zhan }}\\
{\tt\small{mochen.yu, yao.tang, jiajun.liang\}@jiiov.com}}\\
}
\begin{document}
\maketitle
%%%%%%%%% ABSTRACT
\begin{abstract}
Multi-view egocentric hand tracking is a challenging task and plays a critical role in VR interaction. In this report, we present a method that uses multi-view input images and camera extrinsic parameters to estimate both hand shape and pose. To reduce overfitting to the camera layout, we apply crop jittering and extrinsic parameter noise augmentation. Additionally, we propose an offline neural smoothing post-processing method to further improve the accuracy of hand position and pose. Our method achieves 13.92mm MPJPE on the Umetrack dataset and 21.66mm MPJPE on the HOT3D dataset.

\end{abstract}

%%%%%%%%% BODY TEXT
\section{Introduction}
We primarily engage with the world through our hands, experiencing it from a first-person (egocentric) perspective. This egocentric viewpoint is essential for AR/VR applications, where an immersive and intuitive interaction is critical. Commercial virtual reality headsets often utilize multiple cameras for hand tracking, allowing for a larger interaction area. In regions where the cameras' fields of view overlap, multi-view images of the hands can be captured, providing richer 3D data. However, near the boundaries of the interaction area, only single-view images are available. Therefore, it is crucial for hand tracking systems to effectively handle both multi-view and single-view scenarios to ensure accurate and seamless interaction.  Both research and industry\cite{fan2024benchmarks} \cite{fan2023arctic}  \cite{Zhou_2024_CVPR} \cite{zhou20231st} \cite{Ohkawa_2023_CVPR} \cite{banerjee2024introducing} \cite{han2022umetrack} have shown growing interest in improving multi-view egocentric hand tracking performance.

We design an architecture including a feature extracting component, a feature fusion module and several regression parts to process both the input types. Specifically, images are fed into the feature extractor and the feature lifting module separately to obtain 3d features. For single-view input, the features are passed directly to the regression heads to estimate hand pose and shape. For multi-view input, the features are first fused using the feature transform layer (FTL\cite{remelli2020lightweight}) module before continuing through the remaining regressions.

We observe that 2D landmark estimation often outperforms 3D estimations, such as hand position and pose. Based on this observation, we propose an auxiliary post-processing method to optimize 3D estimations.

In summary, our work consists of two key parts: a unified architecture capable of handling both single-view and multi-view inputs, and an offline post-processing step that further improves hand tracking performance. More details will be explained in the following section.

\begin{figure}[t]
  \centering
\includegraphics[width=0.48\textwidth]{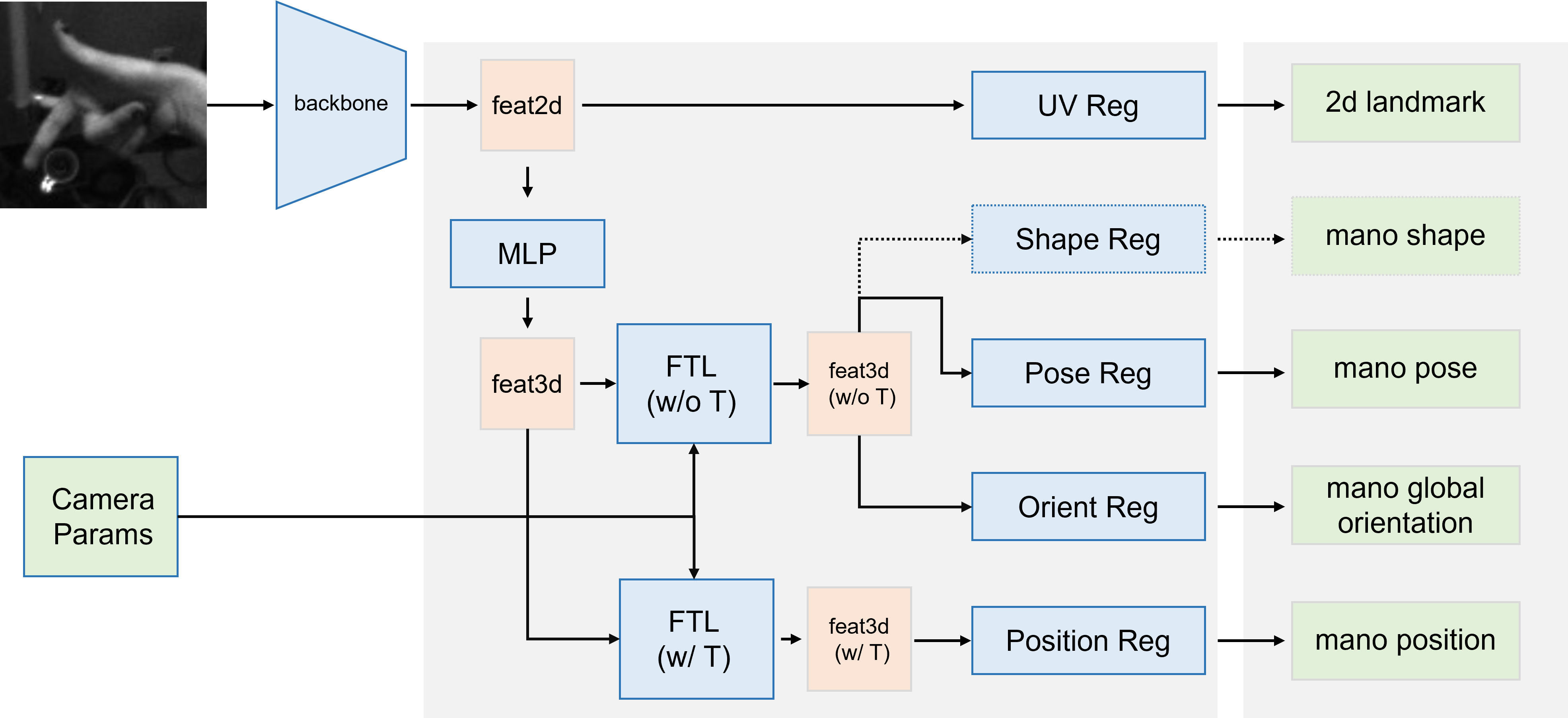}
  \caption{Our method is flexible and supports both monocular and multi-view inputs. For simplicity, the figure illustrates the monocular process. After feature extraction by the backbone, 2D features (feat2d) are obtained and fed into the UV regressor to generate 2D points. Simultaneously, feat2d is processed through an MLP to produce 3D features (feat3d). We then use the FTL module proposed in \cite{han2022umetrack} to generate translation-invariant features (feat3d w/o T) for predicting shape, pose, and global orientation, and translation-aware features (feat3d w/ T) for predicting position.}
  \label{fig:model_structure}
\end{figure}

\section{Method}
\label{Method}
\textbf{Datasets and Preprocessing} In our approach, we exclusively utilized the Umetrack\cite{han2022umetrack} and HOT3D\cite{hot3d} datasets provided for the competition. Preprocessing involved the use of the hand\_tracking\_toolkit\cite{han2022umetrack} to perform perspective cropping. To increase viewpoint diversity, we applied perturbations to the field of view (FOV) and rotation parameters based on the provided perspective crop camera parameters. Additionally, to facilitate the training process for both single-hand and hand-hand interaction scenarios, all left-hand data were flipped to simulate right-hand data.

Our data augmentation strategy included techniques such as crop jittering, random brightness and contrast adjustments, blurring, and Gaussian noise. To improve generalization to different extrinsic camera parameters, we also added random noise in the range of  ($-0.5$, $0.5$) to the translation component of the camera extrinsics.

\textbf{Model Structure} 
Our Model is highly flexible, supporting both monocular and multi-view inputs. The Fig\ref{fig:model_structure}  illustrates the monocular process for simplicity: after feature extraction by the backbone, 2D features (feat2d) are obtained and passed to the UV regressor to predict 2D points. Simultaneously, feat2d is processed through an MLP to generate 3D features (feat3d).

Due to the significant influence of large variations in the translation component(T) of the extrinsic parameters on model predictions, we have decomposed the FTL module into two parts: translation-invariant FTL and translation-variant FTL. The FTL module produces both translation-invariant features (feat3d w/o T) and translation-variant features (feat3d w/ T). The feat3d w/o T is used for predicting shape, pose, and global orientation, while feat3d w/ T is used for position prediction.

For multi-view input, after obtaining feat3d w/o T and feat3d w/ T from each view, we employ a concatenation operation to fuse them into a unified multi-view feat3d w/o T and feat3d w/ T. These fused features are then passed through subsequent modules to generate the final predictions.

\begin{figure}[t]
  \centering
\includegraphics[width=0.48\textwidth]{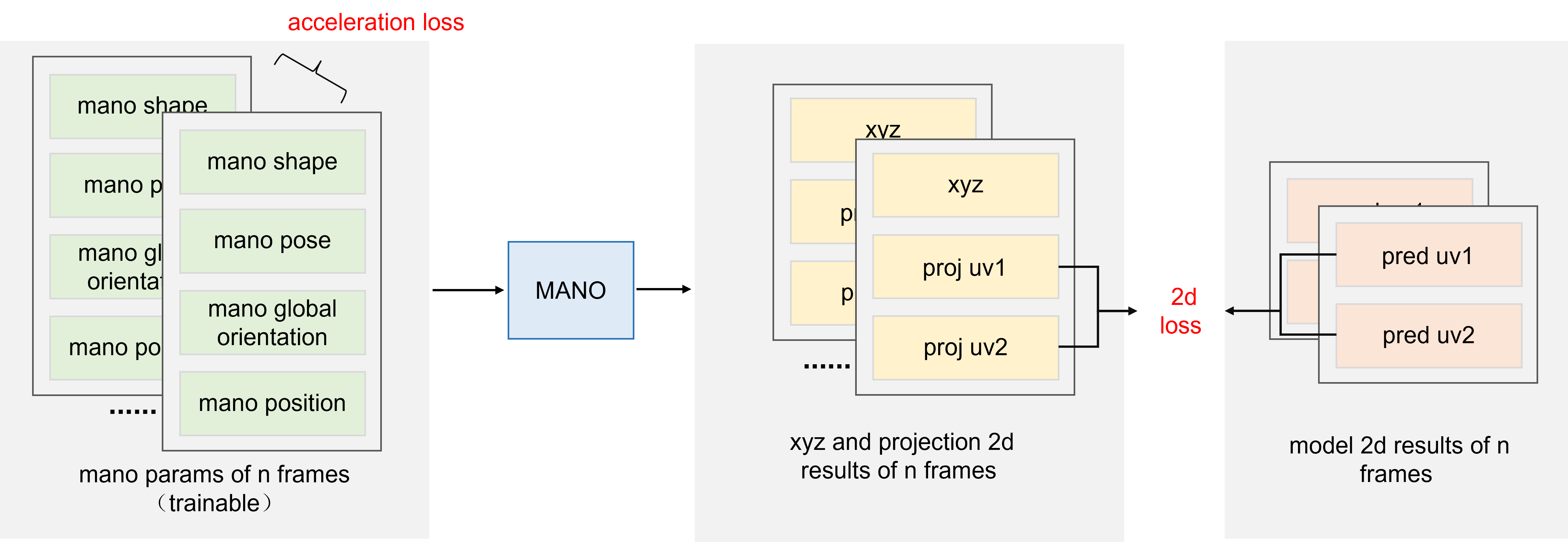}
  \caption{To maintain temporal consistency, we optimize the model's predictions over a sequence as trainable parameters. Simultaneously, we tackle the issue of inaccurate positioning caused by poor generalization to varying extrinsic parameters by employing a 2D projection loss and acceleration loss.}
  \label{fig:neural_smooth}
\end{figure}

\textbf{Neural Smooth} 
Since 2D points can be derived directly from image data with less ambiguity, models typically generalize better when predicting 2D points than 3d predictions. In fact, we observed a significant distribution gap in camera parameters, particularly in translation, between the Umetrack and HOT3D datasets. These discrepancies introduce challenges in accurately predicting positions. To address these generalization issues and leverage temporal information, we propose a method called Neural Smooth.

We begin by obtaining 2D landmarks, as well as predictions for MANO shape, pose, global orientation, and position for each frame. These independent frame-wise predictions form a time series representing hand movements, referred to as the MANO \cite{MANO} parameters. Using these predictions as initial values, we compute the corresponding 3D joints (xyz) and projected 2D points (proj uv) through the MANO model. 2D loss is then calculated between the projected 2D points and the model's predicted 2D results, refining the MANO parameters to improve 2D/3D consistency.

To further ensure temporal consistency, we introduce an acceleration error loss, which promotes smooth transitions between frames and reduces jitter in the predicted hand movements.

Through iterative optimization, Neural Smooth enhances the model's ability to generalize across varying camera parameters, thereby improving the accuracy of position predictions.

The final loss function in Neural Smooth can be formulated as follows:

\[
\mathcal{L} = \lambda_{1} \mathcal{L}_{\text{acce\_pose}} + \lambda_{2} \mathcal{L}_{\text{acce\_orients}} + 
 \lambda_{3} \mathcal{L}_{\text{acce\_position}} + \lambda_{4} \mathcal{L}_{\text{2D}}
\]

where
\[
\mathcal{L}_{\text{acce\_pose}} =  \frac{1}{N-2} \sum_{t=3}^{N} \left| Pose_t - 2Pose_{t-1} + Pose_{t-2} \right|
\]

The $ \mathcal{L}_{\text{acce\_orients}} $ and $ \mathcal{L}_{\text{acce\_position}} $ can also be written in a similar form. In our experiments, $ \lambda_{1} $ to $ \lambda_{4} $ are set to 0.5, 0.5, 0.5, and 1, respectively.

\section{Experiments}
In terms of model architecture, we selected Hiera-base\cite{ryali2023hiera} as the backbone. For training, we used binocular data with an image resolution of 224x224. The training was conducted on 8 NVIDIA 2080TI GPUs, with each GPU processing 16 images, resulting in a total batch size of 128. Each epoch consisted of 256 iterations, and the model was trained for a total of 300 epochs. We use AdamW\cite{loshchilov2017decoupled} optimizer with an initial learning rate of 1e-4 scheduled by cosine scheduler and weight decay 1e-2. 

After obtaining the initial results from the model, we apply the Neural Smooth method for optimization. We utilize the AdamW optimizer with cosine decay, setting the learning rate to 1e-2 and the maximum number of iterations to 500.

\autoref{table1} shows several methods that significantly improved performance in our pose estimation for this challenge.

\begin{table}[h!]
\begin{center}

\begin{tabular}{@{}c|c|c|c@{}}

\hline
ID & method & Umetrack & HOT3d\\
\hline
1 & Baseline & 23.82 & 184.27\\
2 & \multicolumn{1}{l|}{ID1 + extrinsic perturbations} & 17.91 & 36.93 \\
3 & \multicolumn{1}{l|}{ID2 + neural smooth} & 13.92 & 21.66\\
\hline
\end{tabular}
\end{center}
\caption{MPJPE(mm) results on Umetrack/HOT3D test set.}
\label{table1}
\end{table}

For shape estimation, we did not perform any additional optimizations. We simply added a shape prediction module to the aforementioned model \autoref{fig:model_structure} to generate the corresponding predictions.

\section{Conclusion}
In this report, we introduced a flexible model architecture capable of handling both monocular and multi-view inputs, demonstrating its versatility in various hand-tracking scenarios. Through effective data augmentation techniques, such as extrinsic perturbations, and the application of our offline Neural Smooth post-processing method, we significantly improved performance across different datasets. These strategies allowed our model to generalize well to varying camera setups, achieving substantial gains in accuracy, as evidenced by the results on both the Umetrack and HOT3D datasets.

Looking ahead, one promising direction for future work is transitioning Neural Smooth from an offline to an online process, enabling real-time temporal optimization. This advancement would further enhance hand-tracking performance in dynamic and real-world AR/VR applications.

%-------------------------------------------------------------------------

{\small
\bibliographystyle{ieee_fullname}
\bibliography{egbib}
}

\end{document}